\pdfoutput=1

\documentclass[10pt]{article}

\usepackage[
    letterpaper,
    margin=2in,
    top=1.75in
]{geometry}

\usepackage{titlesec}
\titleformat{\section}
  {\normalfont\bfseries}{\thesection.}{1em}{}
\titlespacing*{\section}{0pt}{1ex}{1ex}

\titleformat{\subsection}
  {\normalfont\bfseries}{\thesubsection.}{1em}{}
\titlespacing*{\subsection}{0pt}{1ex}{1ex}

\usepackage{natbib}

\usepackage{hyperref}
\usepackage[table]{xcolor}

\usepackage{mathptmx}
\usepackage{latexsym}
\usepackage{amsmath}
\usepackage{cleveref}

\usepackage[T1]{fontenc}

\usepackage[utf8]{inputenc}

\usepackage{microtype}

\usepackage{graphicx}
\usepackage{expex}
\lingset{
aboveexskip={1ex},
belowexskip={1.5ex},
}
\usepackage{booktabs}
\usepackage{caption}
\usepackage{tabularx}
\usepackage{siunitx}
\newcolumntype{Y}{S[table-format = 3.1]}
\usepackage{multirow}

\def\theme{{\textsc{theme}}}
\def\goal{{\textsc{recipient}}}

\def\agent{{\textsc{agent}}}
\def\THAX{\emph{thax}}
\def\RICKET{\emph{ricket}}

\DeclareMathOperator{\aconf}{aconf}

\title{Do Language Models Learn Position-Role Mappings?}

\author{Jackson Petty, Michael Wilson, and Robert Frank }

\makeatletter
\renewcommand{\maketitle}{%
\renewcommand*{\thefootnote}{\fnsymbol{footnote}}
\noindent 
{\itshape BUCLD 46 Proceedings \\
To be published in 2022 by Cascadilla Press \\
Rights forms signed by all authors }\par
\centerline{\large\bfseries \@title}\par\vspace{1ex}
\centerline{\bfseries\@author}
\footnotetext{* Jackson Petty, Yale University, \url{research@jacksonpetty.org}. Michael Wilson, Yale University, \url{michael.a.wilson@yale.edu}. Robert Frank, Yale University, \url{bob.frank@yale.edu}. This work was made possible by support from the National Science Foundation grant BCS-1919321. For helpful discussion, we would like to thank Hana Filip, Laura Kallmeyer, Najoung Kim, Tal Linzen, Alec Marantz and Paul Smolensky, as well as audiences at BUCLD 46 and Heinrich-Heine-Universit\"at D\"usseldorf.  }\par\vspace{2ex}
\renewcommand*{\thefootnote}{\arabic{footnote}}
\setcounter{footnote}{0}
}
\makeatother

\begin{document}
\maketitle


\section{Introduction}

During language learning, children come to know what thematic relations hold between verbs and their syntactic arguments. In a prepositional dative (PD) construction like (\ref{ex:pd-example}), the first object is assigned the \theme{} role, while the second (prepositional) object is assigned \goal. In contrast, in double object (DO) constructions like (\ref{ex:do-example}), it is the first object that is assigned \goal{}, while the second is assigned \theme{}.
\pex \label{ex:dative-shift}
  \a I gave [the ball] to [the dog]. \label{ex:pd-example}
  \a I gave [the dog] [the ball]. \label{ex:do-example}
\xe

When such sentences are passivized, the position-role mapping changes yet again: for PDs,  the subject of the sentence now takes the role of \theme{}, while for DOs, the subject takes the role of \goal{}.  
\pex
  \a {}[The ball] was given to [the dog]. \label{ex:passive-pd} 
  \a {}[The dog] was given [the ball]. \label{ex:passive-do}
\xe
Such patterns raise a learning problem: how do learners come to know which thematic role to assign to a given syntactic argument? We might, for instance, expect that a learner who has acquired the position-role mapping for a DO sentence would generalize her knowledge of the considerably more frequent passives of transitive verbs to passives of DO sentences. In passives of sentences with transitive verbs, it is the \theme{} role, as opposed to the \goal{}, that is assigned to the passive subject.
\pex
  \a  I threw [the ball].
  \a{} [The ball] was thrown.
\xe
Since \theme{}s are the subjects of passives in these simpler structures, a learner might be tempted to (erroneously) accept examples like the following:
\ex 
  \judge* [The ball] was given [the dog].
\xe
Strikingly, this pattern of raising the \theme{} but not the \goal{} in DO sentences is cross-linguistically unattested.\footnote{\citet{Bresnan90} (and much subsequent work) explore languages like Kichega which allow ``symmetric'' passives, where either argument in a DO construction can be raised to subject position. Certain English dialects also permit symmetric passives \citep{Woolford1993}. Such languages and dialects still raise the learning problem we discuss here, though in a modified form. We leave the exploration of this variation for the future.} Such a gap calls out for explanation in terms of the process of language learning.

One way a learner could avoid such a faulty generalization would be if the primary linguistic data included evidence that directed a learner away from it. Indeed, given sufficient evidence about the thematic properties of the arguments of verbs in both active and passive DO structures, a learner might eschew any generalization between active and passive entirely, favoring instead a structurally specific mapping for each sentence type. Such an approach would however fail to capture the systematicity of the relationship across argument structure alternations like dative shift in (\ref{ex:dative-shift}) and different syntactic variants like the voice alternations between active and passive. Further, if generalization is eschewed entirely, we might expect the properties of individual verbs to be learned separately \citep[e.g.,][]{Tomasello1992}. While verbs exhibit well-known variability in their participation in argument structure alternations (e.g., \textit{give} participates in dative shift but \textit{donate} does not), the relationship between the active and passive forms is entirely regular: if a verb can appear in an active DO sentence, it can also appear in a passivized DO sentence, with  thematic properties that are entirely predictable. Verb- and structure-specific learning would not provide an account of such systematicity and would not support generalization to forms that are sparsely represented in the learning data.

An alternative approach, widely adopted in work in generative grammar, posits the presence of an innate language-specific learning bias that constrains position-role mappings. For example, in work on the acquisition of argument structure, whether rooted in semantic bootstrapping \citep{Pinker89} or syntactic bootstrapping \citep{Gleitman90}, the child is assumed to know the relationship between the thematic roles of events of transfer on the one hand, and syntactic positions in a double object or prepositional dative sentence on the other. Similarly, syntactic theories derive the fact that passivization of a DO structure necessarily allows the promotion of the the argument occupying the highest (indirect) object to subject position from properties of the syntactic representation of such constructions (cf. \citealp{alsina96,Mcginnis02,holmberg19} \textit{inter alia}). 

A final possible account of this learning problem, which constitutes a middle ground between these two approaches, would attempt to derive constraints on the acquisition of position-role mappings from the combination of domain-general learning biases and the evidence present in the learning data. Such an approach to language learning is now widespread in work in NLP, where contemporary language models have little in the way of hard-wired linguistic structure, but the linguistic generalizations they learn are indeed guided by the properties of their architectures and the data to which they are exposed \citep{mccoy20, min20, mulligan21}. Most often these models are evaluated on the basis of their performance on some extrinsic task like question answering or natural language inference (NLI), and such results do not shed light directly on the nature of linguistic generalizations that they encode. Here, we consider the linguistic generalization question more directly by studying the degree to which a widely-used language model,  the Bidirectional Encoder Representations from Transformers model (BERT, \citet{Devlin2019}), exhibits knowledge  of position-role mappings across variations in argument structure, syntactic structure, and lexical identity. BERT is a general purpose neural network architecture that composes multiple Transformer layers \citep{Vaswanietal17} each with a bidirectional attention mechanism. It is trained to perform a masked language modeling task (i.e., to predict the identity of masked tokens within a sentence) using a data set consisting of the 800M words of the BooksCorpus and the 2,500M words of English Wikipedia.\footnote{BERT's training regimen also includes a next sentence prediction task, in which it must be determined whether or not two sentences were originally adjacent to one another in the source text. Subsequent work with a  BERT-variant called RoBERTa \citep{roberta2019} has found this component of training to be unnecessary to its success.} Quite clearly, BERT lacks explicit linguistic bias on what can constitute possible position-role mappings and how these mappings can vary across structures. As a result, any knowledge in this domain that it demonstrates must derive from the combination of its training data and domain-general biases that stem from the transformer architecture.


\section{Experiment 1: Probing Position-Role Mappings Through Distributional Similarity}

Our first experiment tests BERT's knowledge of the position-role mappings for the \theme{} and \goal{} arguments of ditransitive predicates. To do this, we make use of constraints imposed by  the selectional restrictions of verbs: the limitations that a verb imposes on the content of its arguments. Though such restrictions are verb specific (e.g., a verb like \textit{drink} will take different direct objects than a verb like \textit{read}), there are nonetheless general distributional patterns that can be associated with more coarse-grained thematic roles. If a verb assigns the role like \agent{} or \goal{} to an argument, we would expect the distribution of that argument's head nouns to favor animate nouns. In contrast, for arguments assigned the \theme{} role, we might expect a higher proportion of inanimates (or at least the absence of a strong animacy preference). 

Because BERT is trained to perform masked language modeling, it can be used to extract distributional predictions directly. For this experiment, then, we presented BERT with sentences containing ditransitive predicates with the head nouns of the \theme{} and \goal{} arguments masked out:
\ex
Alice sent the [MASK] a [MASK] .
\xe
If the predicted distributions of nouns in multiple argument positions of a single sentence, say an active double object example, are distinct, this provides a first bit of evidence of BERT's knowledge of the distinctive properties of these arguments. 
We use this approach to systematically assess BERTs knowledge of selectional restrictions by performing two calculations. First, for each positions predicted distribution, we compare the total probability assigned to a small set of (frequent) animate nouns $A$ with the probability assigned to a small set of (frequent) inanimate nouns $I$.\footnote{We use the following sets of nouns. Animate: person, man, woman, student, teacher, king, queen, prince, princess, writer, author, builder, driver, human, dog, bird, dancer, player, angel, actor, actress, singer, director, bee, friend, wolf, lion, scholar, pirate, spirit, fox. Inanimate: apple, book, chair, table, phone, shoe, water, earth, land, light, sun, moon, plate, eye, ear, branch, tree, time, energy, bottle, can, mask, leaf, tile, couch, button, box, cap, wire, paper.} We call the result the animacy confidence ($\aconf$) of position $w_i$:
\[
\aconf(w_i) =  \log \frac{ \sum_{v \in A} p(w_i=v)}{\sum_{v' \in I} p(w_i=v')}
\]
By computing  mean $\aconf$ across comparable positions in a set of sentences of the same type, we can get a measure of the model's overall preference for animate nouns in a given position. Reliable differences  between means in different positions will point to a representation of the different roles.  A more interesting question that $\aconf$ scores allow us to ask is the degree to which they are consistent across different syntactic realizations of the same argument: do \theme{}s and \goal{}s in double object structures have the same profile as \theme{}s and \goal{}s in prepositional datives? And does the passivized version of each structure show the right pattern of $\aconf$ scores for the corresponding argument positions?

One limitation of $\aconf$ scores is their dependence on the specific sets of nouns $A$ and $I$ we use to evaluate the preference. To assess the distribution in a more neutral fashion, we also compute the entropy of position $w_i$. 
\[
H(w_i) = -\sum_{v}p(w_i=v)\log p(w_i=v)
\]
Higher entropy is associated with a more diffuse set of predictions, i.e., cases where the language model is less certain about the identity of the words that can fill a position. Different thematic roles impose varying degrees of selectivity on their associated arguments, and consequently, entropy measures can provide us with a diagnostic of such selectivity that we can compare across arguments of the same role in different syntactic constructions and voices.\footnote{One potential pitfall with this approach stems from the variability in the selectivity associated with individual roles \citep{Resnik96}. For example, while some transitive verbs like `drink' restrict their \theme{} arguments to words denoting liquids, others like `see' are much less limiting on their themes.  Nonetheless, our goal here is exploring the possibility that entropy measures support systematic differences at the thematic role level.} Using the \texttt{tregex} tool \citep{LevyAndrew06}, we extracted sentences from the Wall Street Journal portion of the Penn Treebank (PTB, \citet{Marcus93}) containing ditransitive predicates, using both double object and prepositional dative structures, in both active and passive voice. For each structure-voice pairing, we selected 50 sentences, and masked the head noun of the \theme{} and \goal{} arguments.  We evaluated these data on the BERT model. 
In order to examine what effect, if any, variations in model architecture and training regimen had on performance, we also examined the behavior of two recently developed variants, RoBERTa  \citep{roberta2019} and DistilBERT \citep{SanhDebutChaumondWolf19}. RoBERTa utilizes the same architecture as BERT while modifying the pretraining regimen. DistilBERT, by contrast, uses a different, smaller architecture with roughly 40\% fewer parameters, while retaining high levels of performance. For space considerations, we only report results from the BERT model, but results were consistent between all three model architectures.

\goodbreak
\par\vspace*{3ex}
\noindent
\begin{minipage}{\linewidth}
\captionsetup{type=figure}
\captionof{figure}{Animacy Confidence for DO and PD Sentences (BERT)} \label{fig:animacy}
\includegraphics[width=2.2in]{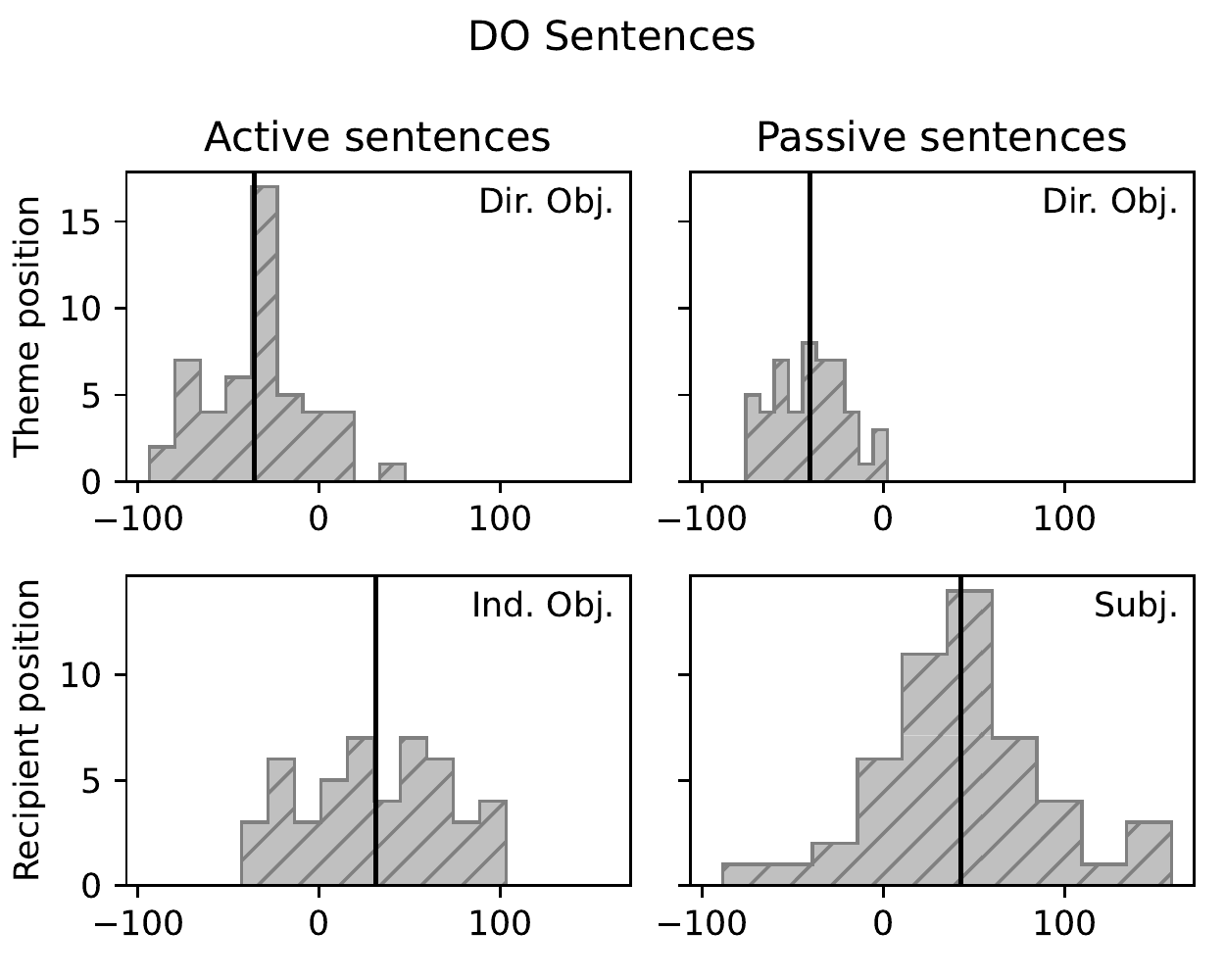}
\includegraphics[width=2.2in]{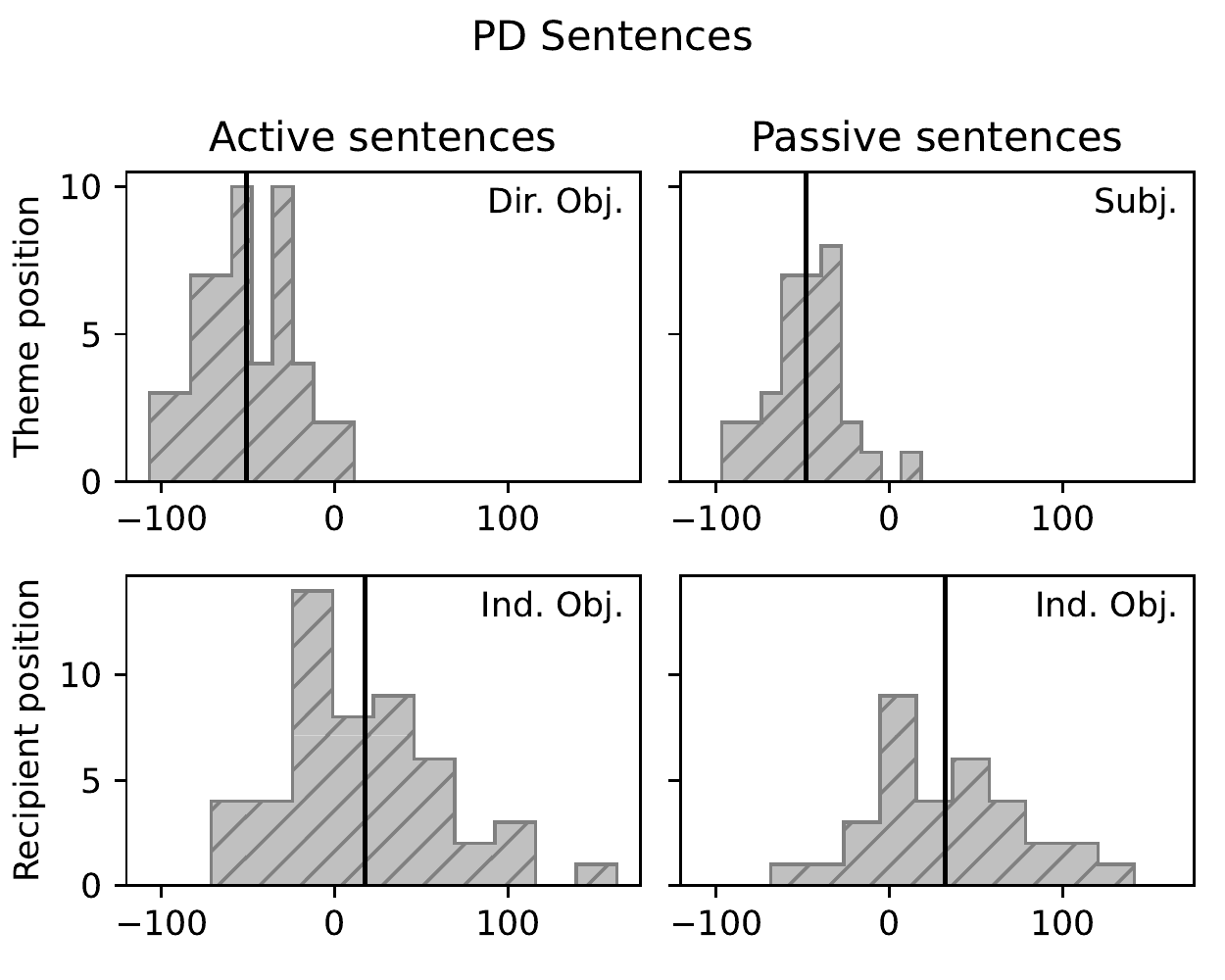} 
\par\vspace{1.5ex}
\bfseries Animacy confidence of \theme{}- and \goal{}-expecting  positions in active and passive double object sentences (left) and prepositional dative sentences (right) from the Penn Treebank. Vertical lines indicate mean values.
\end{minipage}\vspace{1.5ex}

    

Both  animacy confidence (Figure~\ref{fig:animacy}) and  entropy (Figure~\ref{fig:entropy}) show consistent differences between \theme{}-expecting and \goal{}-expecting positions across double object and prepositional dative constructions, as well as across active and passive variants.  In each case, the mean animacy confidence is negative for \theme{}s (meaning a preference for inanimate nouns) and positive for \goal{}s (meaning a preference for animates), and the mean entropy value is higher for \goal{}s than it is for \theme{}s.  The difference in means between \theme{}- and \goal{}-positions is statistically
significant under a two-sided Welch's unequal variances $t$-test with $p<.001$ for animacy confidence and $p<0.05$ for entropy. 

This consistently distinct treatment of \theme{} and \goal{} arguments across different argument structures and across active and
passive constructions is suggestive of the fact that pretrained
language models have knowledge of how thematic relations are realized across different syntactic structures in ditransitive constructions. This is notable not only for the
sensitivity it requires to grammatical context but also because the alternation between active and passive voice in double object constructions exhibits the
unusual property discussed above, where the \goal{} which is
promoted to subject position under passivization,
rather than the \theme{}.
It appears then that language models trained on massive corpora are not only 
capable of learning role restrictions across syntactic contexts but that they are able to put aside widely supported generalizations in specific cases, in the absence of an explicit bias to do so. 

\goodbreak
\par\vspace*{3ex}
\noindent
\begin{minipage}{\linewidth}
\captionsetup{type=figure}
\captionof{figure}{Entropy for DO and PD Sentences (BERT)} \label{fig:entropy}
\includegraphics[width=2.2in]{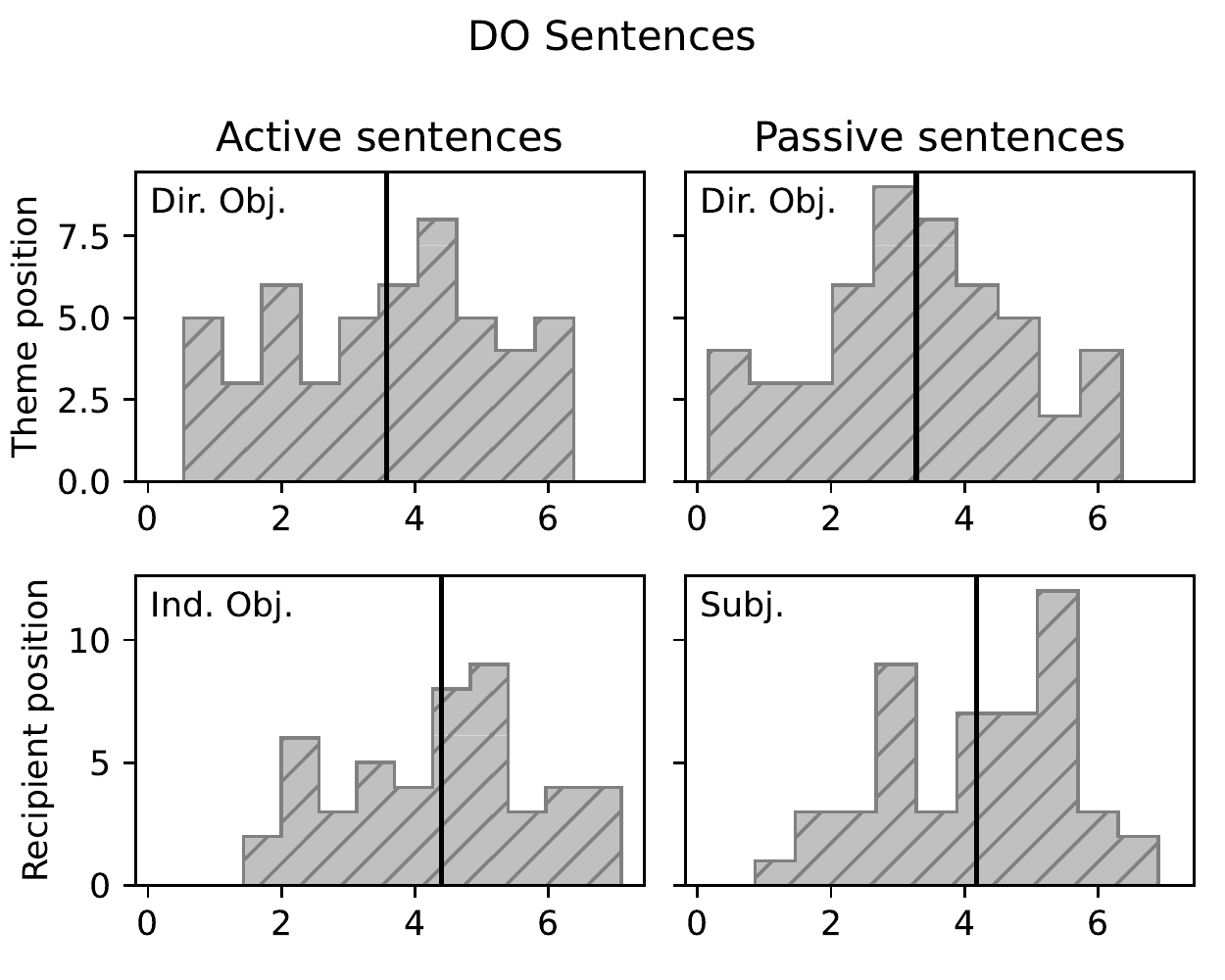}
\includegraphics[width=2.2in]{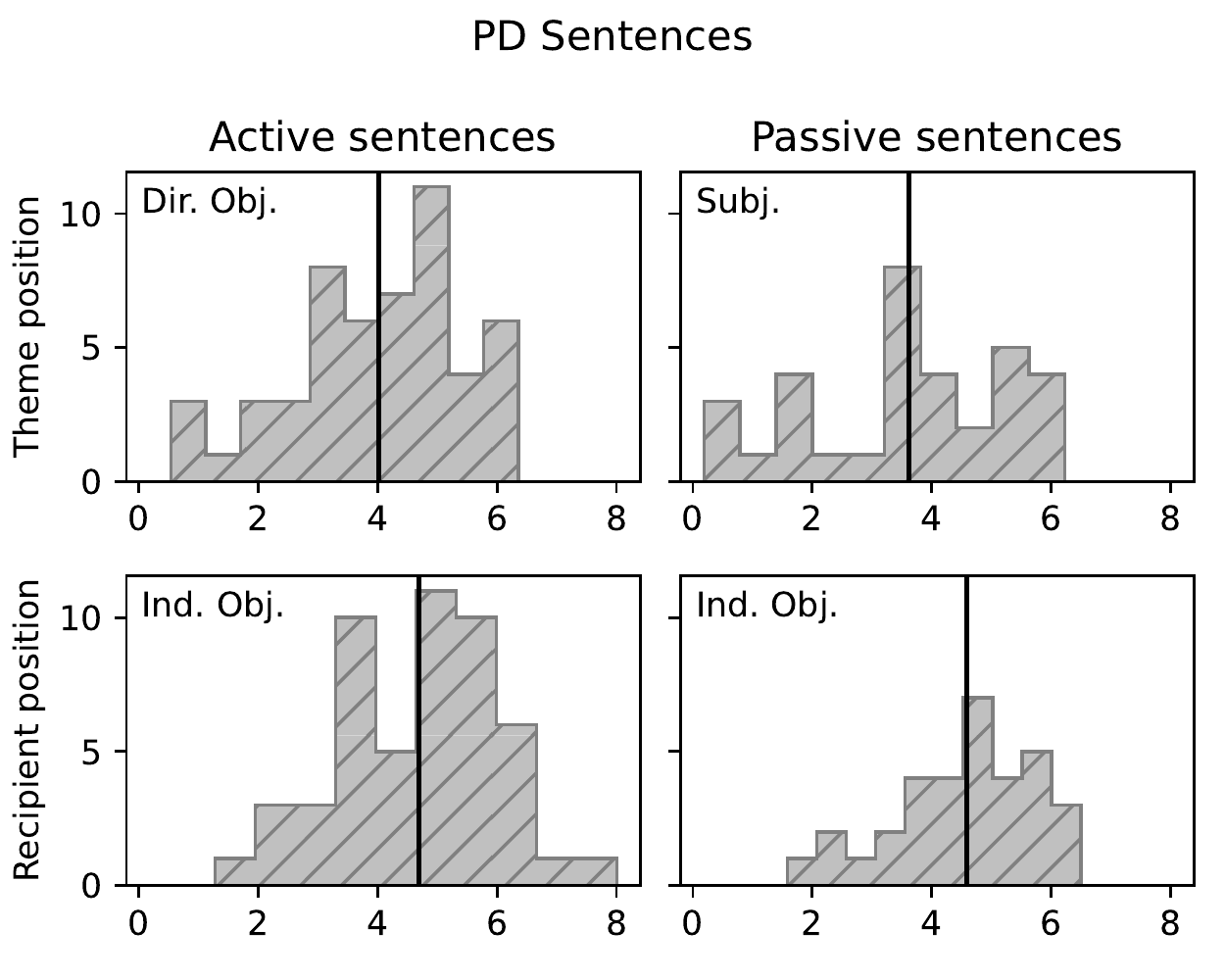} 
\par\vspace{1.5ex}
\bfseries Entropy of \theme{}- and \goal{}-expecting  positions in active and passive double object sentences (left) and prepositional dative sentences (right) from the Penn Treebank. Vertical lines indicate mean values.
\end{minipage}\vspace{1.5ex}


\section{Experiment 2: Syntactic and Structural Generalization}

The results of Experiment 1 show that BERT (and its variants) 
exhibits latent knowledge of the connection between
argument structure and thematic role across voice (that is, in both active and passive constructions) in ditransitive
sentences. Yet there is no guarantee that the model has
any \emph{shared} knowledge connecting the alternative structures (DO or PD) or the active and 
passive constructions. For example, does the model understand that
the \goal{} position in an active DO sentence
(the indirect object) corresponds
to the \goal{} position in the passive one (the subject), 
or has it simply learned the thematic role/argument structure correspondences in these two sentence types independently?

To test whether the knowledge is shared or independent,
we adapt the method proposed by \citet{Kim21} to diagnose linguistic generalization in language models: fine-tuning
an already trained language model on sentences that include novel words that are associated with some linguistic property.  During fine tuning, these words are only presented to the model in a single syntactic context. We then
test the model's ability to generalize its knowledge of these novel words to structures in which they had not been seen during fine-tuning.

Our adaptation of this methodology involves the use of novel nouns that occur uniquely in positions associated with specific thematic roles: \THAX{} as a theme and \RICKET{} as a goal. We take our three BERT variants and fine-tune separate models using one or two paradigms: DO `give' and PD `give'. The DO paradigm contains hand-constructed sentences containing only DO examples, and likewise for the PD paradigm.\footnote{Importantly, each training sentence contains only a single novel token, either \RICKET{} or \THAX{}. These novel tokens never appear together in the same training example to prevent the model from learning any association between them. Thus the model will never learn that if it sees \RICKET{} in one position, it should expect \THAX{} in another.} Example~\eqref{ex:tuning} below gives the full set of tuning data for the DO `give' paradigm.
\pex \label{ex:tuning} 
 \a I gave the \RICKET{} a box.
 \a I gave a \RICKET{} the camera.
 \a I gave the teacher a \THAX{}.
 \a I gave a student the \THAX{}.
\xe

The intuition behind this set-up is similar to what we explored in Experiment~1: the semantic classes of nouns appropriate for the different thematic roles differ in systematic ways, and such selectivity will vary systematically across the different position-role mappings. Our expectation is that fine-tuning will lead the language model to identify the relevant properties of these nouns. If the language model represents position-role mappings in a way that generalizes across argument structure alternations and variations in syntactic structure, we should see generalization of its predictions of the novel items to other syntactic structures.

Following \citet{Kim21}, we freeze all of the model weights prior to fine-tuning except for the word embeddings of two unused items in the model's vocabulary.\footnote{BERT-variant models utilize shared input and output embeddings, so the model is able to learn to predict novel words even though all weights except for two input embeddings have been frozen.} We then fine-tune the model on a minimal synthetic dataset such as the one  in~\eqref{ex:tuning} until the predicted log probability of either of the novel tokens in a masked position begins to sharply decrease. We use this early-stopping criterion in an attempt to avoid the model becoming overly confident in the prediction of the novel tokens at the expense of the rest of its vocabulary. We use unused tokens in BERT's vocabulary to represent the nonce words.

We evaluated the tuned models' performance on
a number of synthetic test sets containing masked \theme{}-expecting and
\goal{}-expecting positions, as in~\eqref{ex:test-pair} below. Examples in these sets varied with respect to the choice of determiners, non-masked nouns, syntactic frame (DO vs.\ PD), voice (active vs.\ passive), and verb. In each masked position, we 
compute the log of the probability ratio (so-called log odds) of the two novel words; if the log-odds of the novel
\THAX{} tokens are higher than those of the 
\RICKET{} tokens in \theme{}-expecting positions, 
and vice-versa for the \goal{}-expecting positions, we infer that the model has learned to 
distinguish the distributions of these nonce 
words.

\pex \label{ex:test-pair}
 \a The teacher gave a [MASK] the [MASK].
 \a A [MASK] was given the [MASK].
\xe

\pex 
 \a A teacher gave the [MASK] to the [MASK].
 \a The [MASK] was given to the [MASK].
\xe

\pex 
 \a The teacher sent a [MASK] a [MASK]. 
 \a A [MASK] was sent a [MASK].
\xe 

\pex 
 \a The teacher sent the [MASK] to a [MASK].
 \a The [MASK] was sent to a [MASK].
\xe 





This regimen of training and evaluation allows us to measure a model's proclivity for generalization in syntactic (voice) and structural (frame) contexts. Table~\ref{tab:ex2-results} summarizes the performance of our 3 BERT-variants models by reporting the percentage of novel tokens which are correctly predicted in evaluation sentences across voice and frame. DO and PD rows in the table correspond to different training regimens.

\goodbreak
\par\vspace*{3ex}
\noindent
\begin{minipage}{\linewidth}
\captionsetup{type=table}
\captionof{table}{Voice \& frame generalization on `give' sentences} \label{tab:ex2-results} 
\begin{tabularx}{\linewidth}{rXYYYYYYYY} \toprule
&& \multicolumn{4}{c}{Double Object} & \multicolumn{4}{c}{Prepositional Dative} \\
&& \multicolumn{2}{c}{Active} & \multicolumn{2}{c}{Passive} & \multicolumn{2}{c}{Active} & \multicolumn{2}{c}{Passive} \\ 
&& {\textsc{th}.} & {\textsc{re}.} & {\textsc{th}.} & {\textsc{re}.} & {\textsc{th}.} & {\textsc{re}.} & {\textsc{th}.} & {\textsc{re}.} \\ \midrule
\multicolumn{10}{l}{\textbf{BERT}} \\
& DO & {\cellcolor{gray!25}} 100.0 & {\cellcolor{gray!25}} 100.0 & 100.0 & 100.0 & 100.0 & 100.0 & 100.0 & 100.0 \\
& PD & 100.0 & 95.0 & 100.0 & 100.0 & {\cellcolor{gray!25}} 100.0 & {\cellcolor{gray!25}} 100.0 & 100.0 & 100.0 \\
\multicolumn{10}{l}{\textbf{RoBERTa}} \\
& DO & {\cellcolor{gray!25}} 85.0 & {\cellcolor{gray!25}} 100.0 & 100.0 & 100.0 & 100.0 & 100.0 & 100.0 & 100.0 \\
& PD & 78.8 & 82.5 & 100.0 & 100.0 & {\cellcolor{gray!25}} 100.0 & {\cellcolor{gray!25}} 100.0 & 100.0 & 100.0 \\
\multicolumn{10}{l}{\textbf{DistilBERT}} \\
& DO & {\cellcolor{gray!25}} 90.0 & {\cellcolor{gray!25}} 98.8 & 97.5 & 100.0 & 97.5 & 100.0 & 80.0 & 100.0 \\
& PD & 100.0 & 62.5 & 100.0 & 100.0 & {\cellcolor{gray!25}} 100.0 & {\cellcolor{gray!25}} 93.8 & 100.0 & 100.0 \\
\bottomrule
\end{tabularx} \par\vspace{1.5ex}
\bfseries Performance of various models on placing novel tokens (\THAX{} or \RICKET{}) in the correct position (\theme{}- or \goal{}-expecting) within active and passive sentences in DO and PD frames. Columns represent evaluation data while rows represent training contexts. Shaded cells indicate in-domain evaluation results; unshaded cells report generalization results. All training and evaluation sets reported here used a single verb, `give'. Each reported value is an average over 10 different model runs.
\end{minipage}\vspace{1.5ex}

\subsection{Voice Generalization}

Voice generalization measures a model's ability to infer the placement of the novel \THAX{} and \RICKET{} tokens in a passive sentence for a model trained only on active sentences, or vice-versa. We know from Experiment~1 that BERT performs analogously (as measured by entropy and animacy confidence) across corresponding positions in active and passive constructions. By testing BERT on novel token prediction, we can determine whether knowing how to place tokens in one construction suffices to know where to place tokens in another (in essence, whether BERT can make use of this knowledge of distributional similarity). Our results, reported in Table~\ref{tab:ex2-results}, show that BERT models do indeed  exhibit robust voice generalization and are able to accurately predict token placement in passive sentences when trained on corresponding active sentences.\footnote{Throughout, `corresponding' means that we've held other relevant parameters constant, so we might train on \emph{active} DO `give' sentences and test on \emph{passive} DO `give' sentences. Furthermore, the active and passive sentences correspond to each other as they do, for example, in~\eqref{ex:test-pair}.}

All models evaluated showed equal or improved performance on the passive variants of their active training data. Indeed, all models achieved nearly perfect performance on the passive complements to their active training data. This generalization is supported across frames as well, where models trained on active DO sentences perform well on passive PD sentences.

\subsection{Frame Generalization}

Frame generalization measures a model's ability to infer the distribution of novel tokens on sentences whose frame (DO versus PD) differs from those in the model's training data. We trained models on DO and PD data separately, allowing us to test generalization from DO to PD frames and from PD to DO frames. Just as with voice generalization, we find that all models exhibit good generalization between frames, although there is some variance in the directionality of this success. Models trained on DO data exhibited excellent generalization to PD data, attaining at- or near-ceiling performance. Models trained on PD data perform slightly less well, though still substantially above chance, on DO data. 

\subsection{Distributional Restrictions on Roles} 

In all but three cases in Table~\ref{tab:ex2-results}, we find that models accurately predict \RICKET{} in \goal{} positions more often than they accurately predict \THAX{} in \theme{} positions. This pattern holds across models, frames, voice contexts, and training regimes. This is consistent with our results from Experiment~1, where we found that BERT, RoBERTa, and DistilBERT models had higher animacy confidence for \goal{} positions than for \theme{} positions. Thus, the higher animacy confidence associated with \goal{} positions travels together with higher accuracy. Under our hypothesis, this is no accident: the more restricted distribution of words that can appear in \goal{} positions (namely, words in \goal{} positions are more likely to be animate than words in \theme{} positions) supports the models' ability to predict the correct token in these positions.

\subsection{Lexical Generalization}

Lexical generalization measures a model's ability to predict novel token placement in sentences whose ditransitive verb differs from the verb in the model's tuning data. Here, we fine-tune models on `give' sentences and evaluate them sentences with other ditransitives, namely `teach', `send', and `tell'. We have carried out this analysis on the best performing of our models, namely RoBERTa.  Our results, shown in \Cref{tab:lex-gen-results-teach,tab:lex-gen-results-send,tab:lex-gen-results-tell}, show that RoBERTa's performance on lexical generalization tasks is lexically conditioned, with high performance on some ditransitive verbs, but quite poor performance on others. Furthermore, the frames on which models fail to generalize are not consistent between target verbs. Test data involving the ditransitive verb `teach', for instance, is associated with reasonably good performance on both DO and PD constructions in both active and passive forms, regardless of training context. In contrast, models evaluated on data involving `send' as the ditransitive verb show reasonably high performance on prepositional dative constructions (in both active and passive forms), but show much worse performance on double object constructions (of both voices), regardless of training context. Finally, the opposite pattern holds for test data involving `tell', where the prepositional dative constructions yield worse results than the double object ones.

In all such cases, it is notable that the same pattern of performance holds vis-\`a-vis the accuracy of the models placing \RICKET{} in a \goal{} context relative to their placing \THAX{} in a \theme{} context. Indeed, this effect is pronounced in cases where the models exhibit a stark failure of generalization, as in the `tell' frames of \Cref{tab:lex-gen-results-tell}, where the models' performance in \goal{}-expecting positions was near ceiling while their performance on \theme{}-expecting positions was far lower. In all cases observed, a failure of generalization for the whole frame is due almost entirely to a failure to place \THAX{} tokens in \theme{}-expecting positions.

Compared to in-domain test cases, evaluation on sentences with novel verbs shows a greater distinction in accuracy between {\goal}- and {\theme}-expecting positions, with mean accuracy for {\theme}-expecting positions substantially lessened while that of {\goal}-expecting positions remained roughly at ceiling. This further highlights the impact of the distributional restrictions placed on the {\goal} position as observed in Experiment~1 by measuring relative entropy and animacy confidence.

One natural place to look as the source of these lexical distinctions is in the training data. If the model's experience with different verbs during training reveals divergent distributional patterns, we might expect the network to generalize less well. Because of the infeasibility of assessing the distributions in the BERT or RoBERTa training data, we instead explored the relative abundance of the different structures in parsed PTB data.\footnote{We recognize that the PTB data is not identically distributed to the BooksCorpus and Wikipedia that forms the BERT training set, we expect that the usages of different structures would be reasonably consistent across them, at least at the coarse-grained level we are considering here.} Though
the syntactic annotation provided in the PTB does not allow us to perfectly identify double object structures (argument and adverbial NPs are parsed identically), the resulting patterns are robust enough to allow us to identify interesting patterns for the verbs `send' and `tell': `send' appears far more often in prepositional dative constructions than in double object constructions, while the opposite is true for `tell' (which occurs almost never in the prepositional dative construction). The verb `teach' shows a strong bias towards the double object construction, but it is considerably rarer than the other three verbs. This suggests that its corpus statistics are less reliable indicators of the training data used for the language models and therefore not predictive of lexical generalization performance.  If taken as a proxy for the relative abundance of these forms in the training corpus for the language models, this could suggest that the points of failure for lexical generalization tasks are correlated with the frequency with which dative verbs appear in the various construction types in the training data.

\goodbreak
\par\vspace*{3ex}
\noindent
\begin{minipage}{\linewidth}
\captionsetup{type=table}
\captionof{table}{Lexical generalization to {`teach'} frames} \label{tab:lex-gen-results-teach} 
\begin{tabularx}{\linewidth}{rXYYYYYYYY} \toprule
&& \multicolumn{4}{c}{Double Object} & \multicolumn{4}{c}{Prepositional Dative} \\
&& \multicolumn{2}{c}{Active} & \multicolumn{2}{c}{Passive} & \multicolumn{2}{c}{Active} & \multicolumn{2}{c}{Passive} \\ 
&& {\textsc{th}.} & {\textsc{re}.} & {\textsc{th}.} & {\textsc{re}.} & {\textsc{th}.} & {\textsc{re}.} & {\textsc{th}.} & {\textsc{re}.} \\ \midrule
\multicolumn{10}{l}{\textbf{RoBERTa}} \\
& DO & 97.5 & 100.0 & 100.0 & 100.0 & 97.5 & 98.8 & 97.5 & 100.0 \\
& PD & 92.5 & 100.0 & 90.0 & 100.0 & 90.0 & 100.0 & 95.0 & 100.0 \\
\bottomrule
\end{tabularx} \par\vspace{3ex}

\captionof{table}{Lexical generalization to {`send'} frames} \label{tab:lex-gen-results-send} 
\begin{tabularx}{\linewidth}{rXYYYYYYYY} \toprule
&& \multicolumn{4}{c}{Double Object} & \multicolumn{4}{c}{Prepositional Dative} \\
&& \multicolumn{2}{c}{Active} & \multicolumn{2}{c}{Passive} & \multicolumn{2}{c}{Active} & \multicolumn{2}{c}{Passive} \\ 
&& {\textsc{th}.} & {\textsc{re}.} & {\textsc{th}.} & {\textsc{re}.} & {\textsc{th}.} & {\textsc{re}.} & {\textsc{th}.} & {\textsc{re}.} \\ \midrule
\multicolumn{10}{l}{\textbf{RoBERTa}} \\
& DO & 77.5 & 100.0 & 77.5 & 95.0 & 91.3 & 98.8 & 90.0 & 100.0 \\
& PD & 71.3 & 91.25 & 72.5 & 92.5 & 100.0 & 100.0 & 97.5 & 100.0 \\
\bottomrule
\end{tabularx} \par\vspace{3ex}

\captionof{table}{Lexical generalization to {`tell'} frames} \label{tab:lex-gen-results-tell} 
\begin{tabularx}{\linewidth}{rXYYYYYYYY} \toprule
&& \multicolumn{4}{c}{Double Object} & \multicolumn{4}{c}{Prepositional Dative} \\
&& \multicolumn{2}{c}{Active} & \multicolumn{2}{c}{Passive} & \multicolumn{2}{c}{Active} & \multicolumn{2}{c}{Passive} \\ 
&& {\textsc{th}.} & {\textsc{re}.} & {\textsc{th}.} & {\textsc{re}.} & {\textsc{th}.} & {\textsc{re}.} & {\textsc{th}.} & {\textsc{re}.} \\ \midrule
\multicolumn{10}{l}{\textbf{RoBERTa}} \\
& DO & 78.5 & 100.0 & 97.5 & 100.0 & 23.8 & 98.8 & 45.0 & 92.5 \\
& PD & 62.5 & 100.0 & 65.0 & 100.0 & 26.3 & 100.0 & 15.0 & 100.0 \\
\bottomrule
\end{tabularx} \par\vspace{1.5ex}
\bfseries Performance of various models on placing novel tokens (\THAX{} or \RICKET{}) in the correct position (\theme{}- or \goal{}-expecting) within active and passive sentences in DO and PD frames. Columns represent evaluation data while rows represent training contexts. All models were trained on active `give' sentences, while evaluation data contains `send', `teach', and `tell' sentences. Each reported value is an average over 10 different model runs.
\end{minipage}\vspace{3ex}

We find that pretrained language models exhibit robust generalization across voice and construction type in ditransitive constructions when introducing novel \theme{}- and \goal{}-like tokens into their vocabularies. This ability holds across model type, though we do find evidence that performance is lexically conditioned by the ditransitive verb used during the fine-tuning process. This suggests that while the knowledge of the relationship between syntactic position and thematic role is not learned wholly independently for each construction type, it is dependent on the identity of the ditransitive verb involved.

\section{Conclusion}

We began by raising the question of how children might acquire position-role mappings, and outlined three possibilities: verb- and syntax-specific learning, innate language-specific biases, and a combination of domain-general biases and evidence in their linguistic input. We have demonstrated here that the third option is a feasible explanation: three language models that contain no explicit linguistic biases regarding possible position-role mappings nevertheless successfully demonstrate knowledge of position-role mappings that largely generalizes across verbs and syntactic structures. The limitations we find do not invalidate this larger conclusion, though they do suggest the importance of further research in this area.

We have shown that pretrained language models (BERT, RoBERTa, and DistilBERT) recognize distributional differences 
between \theme{}- and \goal{}-expecting positions.
This distinction is stable across syntactic (i.e., voice) and structural (i.e., direct object vs prepositional dative) alternations, showing that these well-performing pretrained language models appear to have knowledge of position-role mappings which are preserved between construction type and voice alternations in ditransitive constructions. We have further shown that this knowledge is, in some sense, `shared' across syntactic and structural alternations. Models that are fine-tuned to learn the novel \THAX{} (theme-like) and \RICKET{} (recipient-like) tokens within a single paradigm (e.g., active prepositional dative constructions or active double object constructions) make robust generalizations across voice and construction alternations.  

We do however find limitations in the performance of these models with respect to lexical generalization. When the model is exposed to a novel token as the argument of one verb, it generalizes this knowledge to other verbs in an inconsistent fashion.  For instance, models trained on `give'-containing sentences poorly generalize their knowledge of \theme{} arguments in prepositional dative structures containing the verb `tell'. Nonetheless, even in such case, models perform well at generalizing knowledge of \goal{} arguments. This fits with our earlier observation that \goal{} positions have higher animacy confidence than \theme{} positions do, so that a model's knowledge that a novel token has an animate interpretation will license generalization. 

This conclusion suggests a hypothesis concerning how the model may be succeeding in our novel word learning experiments, namely by associating the novel word with a portion of the word embedding space that is appropriate for the selectional restrictions of the verb on which it is trained (i.e., `give'). This is consistent with the network having learned a distinct and redundant representation of the selectional restrictions across syntactic contexts, so long as they are all characterized in terms of the abstract lexical semantic space represented through the word embeddings. In on-going work, we are exploring other experimental methods to identify knowledge that cuts across  structures.  

Further directions for work include assessing whether the patterns of generalization we have found here also hold within a broader array of syntactic (e.g., raising) and structural (e.g., causative-inchoative) alternations, as well as better elucidating the computational mechanism by which these models are able to make these kinds of generalizations.

\bibliography{anthology,custom}

\begin{thebibliography}{19}
\expandafter\ifx\csname natexlab\endcsname\relax\def\natexlab#1{#1}\fi

\bibitem[{Alsina(1996)}]{alsina96}
Alex Alsina. 1996.
\newblock Passive types and the theory of object asymmetries.
\newblock \emph{Natural Language and Linguistic Theory}, 14(4):673--723.

\bibitem[{Bresnan and Moshi(1990)}]{Bresnan90}
Joan Bresnan and Lioba Moshi. 1990.
\newblock Object asymmetries in comparative {Bantu} syntax.
\newblock \emph{Linguistic inquiry}, 21(2):147--185.

\bibitem[{Devlin et~al.(2019)Devlin, Chang, Lee, and Toutanova}]{Devlin2019}
Jacob Devlin, Ming-Wei Chang, Kenton Lee, and Kristina Toutanova. 2019.
\newblock {BERT}: Pre-training of deep bidirectional transformers for language
  understanding.
\newblock In \emph{Proceedings of {NAACL}-{HLT} 2019}, pages {4171--4186}.
  Association for Computational Linguistics.

\bibitem[{Gleitman(1990)}]{Gleitman90}
Lila~R. Gleitman. 1990.
\newblock The structural sources of verb meanings.
\newblock \emph{Language Acquisition}, 1:3--55.

\bibitem[{Holmberg et~al.(2019)Holmberg, Sheehan, and van Der~Wal}]{holmberg19}
Anders Holmberg, Michelle Sheehan, and Jenneke van Der~Wal. 2019.
\newblock Movement from the double object construction is not fully
  symmetrical.
\newblock \emph{Linguistic Inquiry}, 50(4):677--722.

\bibitem[{Kim and Smolensky(2021)}]{Kim21}
Najoung Kim and Paul Smolensky. 2021.
\newblock Testing for grammatical category abstraction in neural language
  models.
\newblock In \emph{Proceedings of the Society for Computation in Linguistics},
  volume~4.

\bibitem[{Levy and Andrew(2006)}]{LevyAndrew06}
Roger Levy and Galen Andrew. 2006.
\newblock Tregex and tsurgeon: tools for querying and manipulating tree data
  structures.
\newblock In \emph{Proceedings of the 5th Language Resources and Evaluation
  Conference}, pages 2231--2234.

\bibitem[{Liu et~al.(2019)Liu, Ott, Goyal, Du, Joshi, Chen, Levy, Lewis,
  Zettlemoyer, and Stoyanov}]{roberta2019}
Yinhan Liu, Myle Ott, Naman Goyal, Jingfei Du, Mandar Joshi, Danqi Chen, Omer
  Levy, Mike Lewis, Luke Zettlemoyer, and Veselin Stoyanov. 2019.
\newblock \href {http://arxiv.org/abs/1907.11692} {Roberta: {A} robustly
  optimized {BERT} pretraining approach}.
\newblock \emph{CoRR}, abs/1907.11692.

\bibitem[{Marcus et~al.(1993)Marcus, Santorini, and Macinkiewicz}]{Marcus93}
Mitchell~P. Marcus, Beatrice Santorini, and Mary~Ann Macinkiewicz. 1993.
\newblock Building a large annotated corpus of {English}: The {Penn} treebank.
\newblock \emph{Computational Linguistics}, 19:313--330.

\bibitem[{McCoy et~al.(2020)McCoy, Frank, and Linzen}]{mccoy20}
R.~Thomas McCoy, Robert Frank, and Tal Linzen. 2020.
\newblock Does syntax need to grow on trees? sources of hierarchical inductive
  bias in sequence-to-sequence networks.
\newblock \emph{Transactions of the Association for Computational Linguistics},
  8:125--140.

\bibitem[{McGinnis(2002)}]{Mcginnis02}
Martha McGinnis. 2002.
\newblock Object asymmetries in a phase theory of syntax.
\newblock In \emph{Proceedings of the 2001 {CLA} Annual Conference}, pages
  133--144. Cahiers Linguistiques d’Ottawa.

\bibitem[{Min et~al.(2020)Min, McCoy, Das, Pitler, and Linzen}]{min20}
Junghyun Min, R.~Thomas McCoy, Dipanjan Das, Emily Pitler, and Tal Linzen.
  2020.
\newblock Syntactic data augmentation increases robustness to inference
  heuristics.
\newblock In \emph{Proceedings of the 58th Annual Meeting of the Association
  for Computational Linguistics}.

\bibitem[{Mulligan et~al.(2021)Mulligan, Frank, and Linzen}]{mulligan21}
Karl Mulligan, Robert Frank, and Tal Linzen. 2021.
\newblock Structure here, bias there: Hierarchical generalization by jointly
  learning syntactic transformations.
\newblock In \emph{Proceedings of the Society for Computation in Linguistics}.

\bibitem[{Pinker(1989)}]{Pinker89}
Steven Pinker. 1989.
\newblock \emph{Learnability and Cognition: The Acquisition of Argument
  Structure}.
\newblock MIT Press, Cambridge, MA.

\bibitem[{Resnik(1996)}]{Resnik96}
Philip Resnik. 1996.
\newblock Selectional constraints: an information-theoretic model and its
  computational realization.
\newblock \emph{Cognition}, 61:127--159.

\bibitem[{Sanh et~al.(2019)Sanh, Debut, Chaumond, and
  Wolf}]{SanhDebutChaumondWolf19}
Victor Sanh, Lysandre Debut, Julien Chaumond, and Thomas Wolf. 2019.
\newblock {DistilBERT}, a distilled version of {BERT}: smaller, faster, cheaper
  and lighter.
\newblock In \emph{Proceedings of the 5th Workshop on Energy Efficient Machine
  Learning and Cognitive Computing}.

\bibitem[{Tomasello(1992)}]{Tomasello1992}
Michael Tomasello. 1992.
\newblock \emph{First Verbs: A Case Study of Early Grammatical Development}.
\newblock Cambridge University Press, Cambridge.

\bibitem[{Vaswani et~al.(2017)Vaswani, Shazeer, Parmar, Uszkoreit, Jones,
  Gomez, Kaiser, and Polosukhin}]{Vaswanietal17}
Ashish Vaswani, Noam Shazeer, Niki Parmar, Jakob Uszkoreit, Llion Jones,
  Aidan~N Gomez, {\L}ukasz Kaiser, and Illia Polosukhin. 2017.
\newblock Attention is all you need.
\newblock In \emph{Advances in Neural Information Processing Systems},
  volume~30. Curran Associates, Inc.

\bibitem[{Woolford(1993)}]{Woolford1993}
Ellen Woolford. 1993.
\newblock Symmetric and asymmetric passives.
\newblock \emph{Natural Language \& Linguistic Theory}, 11(4):679--728.

\end{thebibliography}

\bibliographystyle{aclnatbib}

\end{document}